\documentclass{article}

\usepackage{microtype}
\usepackage{soul,xcolor}

\def\incrementality{1}

\usepackage{graphicx}
\usepackage{booktabs} 

\newcommand{\ot}[1]{\textcolor{black}{#1}}
\newcommand{\jr}[1]{\textcolor{black}{#1}}

\newcommand{\ccc}[1]{\textcolor{black}{#1}}

\newcommand{\omitme}[1]{}

\usepackage{lscape}
\usepackage{array}
\lccode`0=`0
\lccode`1=`1
\lccode`2=`2
\lccode`3=`3
\lccode`4=`4
\lccode`5=`5
\lccode`6=`6
\lccode`7=`7
\lccode`8=`8
\lccode`9=`9
\newcolumntype{P}[1]{>{\hspace{0pt}}p{#1}}

\usepackage{hyperref}
\usepackage{placeins}
\usepackage{amssymb}
\usepackage{amsmath}
\usepackage{amsthm}

\usepackage{hhline}
\usepackage{url}
\usepackage{rotating}
\usepackage{ifthen}
\usepackage{algorithm}
\usepackage{authblk}
\usepackage{algorithmicx}
\usepackage{algpseudocode}
\usepackage{epsfig}
\usepackage{multirow}
\usepackage{array}
\usepackage{color}
\usepackage{multicol}
\usepackage{graphicx}
\usepackage{caption}
\usepackage{float}
\usepackage{hhline}
\usepackage{calc}
\usepackage{enumitem}
\usepackage{todonotes} 
\usepackage{bold-extra}
\usepackage{footnote}
\makesavenoteenv{tabular}
\makesavenoteenv{table}
\usepackage{threeparttable}
\usepackage{adjustbox}
\usepackage[a4paper, total={6in, 8in}]{geometry}

\usepackage{hyperref}
\hypersetup{
    colorlinks,
    citecolor=black,
    filecolor=black,
    linkcolor=blue,
    urlcolor=black}

\begin{document}

\newcommand{\R}{\mathbb{R}}
\newcommand{\N}{\mathbb{N}}




\title{Fully Parallel Hyperparameter Search: Reshaped Space-Filling}
\author[1]{ M.-L. Cauwet$^+$}
\author[2]{C. Couprie}
\author[3]{J. Dehos}
\author[2]{P. Luc}
\author[2]{J. Rapin}
\author[2]{M. Riviere}
\author[3]{F. Teytaud}
\author[2]{O. Teytaud}
\def\oldauthor{
Marie-Liesse Cauwet,
Camille Couprie,
Jeremy Rapin,
Pauline Luc,
Morgane Riviere,
Olivier Teytaud
}

\affil[1]{ESIEE, Universit\'e Paris-Est, LIGM (UMR 8049), CNRS, ENPC, ESIEE Paris, UPEM, F-77454, Marne-la-Vall\'ee, France}
\affil[2]{Facebook AI Research}
\affil[3]{Universit\'e du Littoral C\^ote d'Opale }
\date{}

\setcounter{tocdepth}{4}
\setcounter{secnumdepth}{4}

\newcommand{\nico}[1]{{\color{blue} N: #1}}




%





\def\longversion{0}
\def\CQFD{\fbox{}}
\def\qed{\fbox{}}
\def\U{{\cal U}}
\def\E{{\mathbb E}}
\def\NN{\mathcal{N}}
\def\Z{\mathbb{Z}}
\def\G{\mathcal{G}}
\def\e{\epsilon }
\def\I{{\cal I}}
\def\P{{\bf P}}
\def\P{{\mathbb P}}
\def\a{\alpha }
\def\d{\delta }
\def\R{{\mathbb R}}
\def\N{{\mathbb N}}
\def\W{\mathcal{W}}
\def\c{{\rm col}}
\def\r{{\rm row}}
\def\w{\bm{w}}
\def\hx{\hat x}
\def\hA{\hat A}
\def\hB{\hat B}
\def\LHS{{\rm LHS}}
\def\D{{\mathcal D}}
\def\disc{{\rm disc}}
\def\disp{{\rm disp}}
\def\sdisp{{\rm sdisp}}

\def \F2F5{F$_2$F$_5$}

\newtheorem{lemma}{Lemma}
\newtheorem{conj}{Conjecture} 
\newtheorem{prop}{Property} 
\newtheorem{cor}{Corollary} 
\newtheorem{theorem}{Theorem} 
\newtheorem{remark}{Remark}
\newtheorem{example}{Example}
\newtheorem{defn}{Definition}
\newtheorem*{refproof}{Proof}

\sloppy


 \newcommand{\cfbox}[2]{%
     \colorlet{currentcolor}{.}%
    {\color{#1}%
     \fbox{\color{currentcolor}#2}}%
 }




\maketitle
\begin{abstract}
 Space-filling designs such as scrambled-Hammersley,
Latin Hypercube Sampling and Jittered Sampling have been proposed for fully parallel hyperparameter search, and were shown to be more effective than random or grid search. In this paper, we show that these designs only improve over random search  by a constant factor. In contrast, we introduce a new approach based on \emph{reshaping} the search distribution, which leads to substantial gains over random search, both theoretically and empirically. We propose two flavors of reshaping.
   First, when the distribution of the optimum is some known $P_0$, we propose Recentering, which uses as search distribution a modified version of $P_0$ tightened closer to the center of the domain, in a dimension-dependent and budget-dependent manner.
Second, we show that in a wide range of experiments with $P_0$ unknown, using a proposed Cauchy transformation, which simultaneously has a heavier tail (for unbounded hyperparameters) and is closer to the boundaries (for bounded hyperparameters), leads to improved performances. {Besides artificial experiments and simple real world tests on clustering or Salmon mappings, we check our proposed methods on expensive artificial intelligence tasks such as attend/infer/repeat, \ot{video next frame segmentation forecasting} and progressive generative adversarial networks.}
\def\poubelle{
\begin{itemize}
\item TODO anonymized github
\item TODO metacalais initialization
\item TODO metacalais results in artif ?
\item TODO DEinit ?
\item TODO better name than calais ?
\item TODO bullshit conclusion
\end{itemize}}\\
$^+$ Main author of the theoretical analysis.
\end{abstract}
\setcounter{tocdepth}{4}
\setcounter{secnumdepth}{4}
\def\a{$\dagger$}
\def\b{$\ddagger$}
\def\c{$|$}
\def\d{$||$}
\def\ee{$+$}
\def\f{$*$}
\def\ob{$\#$}
\begin{table*}[htb]
\centering
\scriptsize
\begin{threeparttable}
\caption{Comparison between non-reshaped sampling methods {for one shot optimization}. Red boxed results are contributions of the paper. The sequences {of samples} are of size $n$ in dimension $d$. $C_d$ is a dim-dependent constant.
 \label{bigtable} }
\begin{tabular}{|c ||c|c|c|c|c|}
    \hhline{-||-----}
    Sequence& Discrepancy \tnote{\ee}~~~~~~~	& Incrementality &  Randomized 	& Stochastic  & Stochastic dispersion  \\ 
    		&  & (see SM)	&		 with PDF$>0$\tnote{\a}~~~& dispersion\tnote{\b}~~~~~~~~ 					& preserved by projection\tnote{\f}~~~\\
    		\hhline{=::=====}
    $\LHS$  & $\Theta(\sqrt{d/n})$   	& \cfbox{red}{$2n$, optimal}     	& yes	& \cfbox{red}{$1/n^{1/d}$}  & \cfbox{red}{yes} \\
     & ~\cite{doerr} & (SM) &  & (Prop. \ref{zeporp}) & \\
    \hhline{-||-----}
    Grid 	& $1/n^{1/d}$  & no &  no                	& $1/n^{1/d}$ (easy)&no (easy)  \\ \hhline{-||-----}
    Jittered & $\sqrt{d \log n}/n^{1/2 + 1/2d}$  & \cfbox{red}{$2^{d}n$, optimal}  & yes&\cfbox{red}{$1/n^{1/d}$}  &\cfbox{red}{yes} \\
     & $\star$ & (SM) & & (Prop. \ref{jitproj}) & (Prop. \ref{jitproj})\\ 
    \hhline{-||-----}	
    Random  & $\log\log(n)^{\frac12}/\sqrt{n}$	& $+1$ & yes 	&  $1/n^{1/d}$ (easy)&yes (easy)\\ 
     & ~\cite{k} & & & & \\
    \hhline{-||-----}
    Halton & $(1+o(1))\times $ & $+1$ & no &   $1/n^{1/d}$ \tnote{\ob}&yes but \tnote{\ob}\\
       & $C_d (\log n)^d / n$  & & & & different constant\\ \hhline{-||-----}
    Hammersley & $(1+o(1))C_d\times$ & \cfbox{red}{not $n+k\log(n)^{d-\e}$}  & no &$1/n^{1/d}$ \tnote{\ob}	&yes but\tnote{\ob}\\ 
    		    & $ (\log n)^{d-1} / n$		&(SM)	$\e \leq 1, k\leq n-1$	&			& & different constant	\\ \hhline{-||-----}
    Scrambled 	& as Halton, & $+1$ 	 & no	&$1/n^{1/d}$ \tnote{\ob}&yes but \tnote{\ob}\\ 
    Halton 	& better constant& 	& 			&  		& different constant\\ \hhline{-||-----}
    Scrambled    & as Hammersley,  & \cfbox{red}{not $n+k\log(n)^{d-\e}$} & no	&   $1/n^{1/d}$ \tnote{\ob} &yes but \tnote{\ob}\\ 
    Hammersley 	& better constant& (SM) $\e \leq 1, k\leq n-1$ 	&  	& 		& different constant\\ \hhline{-||-----}
    Sobol 	& as Halton & $+1$ & no& $\log(n) / n^{1/d}$	&\cfbox{red}{yes but dif. const.}\ \ \ \tnote{\c} \\ \hhline{-||-----}
    Random & as original LDS (up to & as original LDS & yes  &as original LDS (up to&   \\ 
    -Shift LDS & dim-dependent factor) & & & dim-dependent factor)&\\
      \hhline{-||-----}
\end{tabular}
\begin{tablenotes}
\item[\ee] The discrepancy of a projection $\Pi(S)$ of a sequence $S$, when $\Pi$ is a projection to a subset of indices, is less or equal to the discrepancy of $S$. Therefore we do not discuss the stability of the sequence in terms of discrepancy of the projection to a subspace, contrarily to what we do for dispersion (for which significant differences can occur, between $S$ and $\Pi(S)$).
\item[\a] {"randomized with PDF$>0$" means that the sampling is randomized with a probability distribution function (averaged over the sample) strictly positive over all the domain. }
\item[\b] Optimal rate is $O(1/n^{1/d})$ ~\cite{Sukharev}.
~~[\f] We  consider subspaces parallel to axes, i.e. switching to a subset of indices. We request that the dependency in the dimension becomes the dimension of the subspace.
\item[\c] The bound on the distance to the optimum is an immediate application of the discrepancy, and low discrepancy is preserved by switching to a subspace, hence this positive result.
\item[\ob] Constants depend on which variables are in the subspace - first hyperparameters are ``more'' uniform~\cite{bousquet}.~~~~~~~[$\star$] ~\cite{pausinger}
\end{tablenotes}
\end{threeparttable}
\end{table*}
\begin{table*}[t]
    \centering\scriptsize
    \begin{tabular}{|P{0.03\textwidth}|P{0.10\textwidth}|P{0.10\textwidth}|P{0.10\textwidth}|P{0.10\textwidth}|P{0.10\textwidth}|P{0.10\textwidth}|P{0.10\textwidth}|P{0.10\textwidth}|P{0.10\textwidth}|}
\hline
 & 30&100&300&1000&3000&10000&30000&100000&300000\\
\hline
3& Scr Halton & Scr Halton Plus Middle Point & O Rctg1.2 Scr Halton & O Scr Hammersley & Cauchy Rctg.55 Scr Hammersley & Cauchy Rctg.55 Scr Hammersley & Scr Hammersley & O Rctg.7 Scr Halton & Random \\
\hline
18& Scr Halton Plus Middle Point & Scr Halton Plus Middle Point & O Rctg.7 Scr Halton & Scr Halton & Rctg1.2 Scr Hammersley & O Rctg.4 Scr Hammersley & {\bf{Meta Rctg}} & Cauchy Rctg.55 Scr Hammersley & O Rctg.7 Scr Halton \\
\hline
25& {\bf{Meta Rctg}} & Rctg.4 Scr Halton & O Rctg.4 Scr Halton & {\bf{Meta Rctg}} & Rctg.7 Scr Hammersley & Rctg.7 Scr Hammersley & O Rctg.7 Scr Halton & O Rctg.7 Scr Hammersley & Rctg.7 Scr Halton \\
\hline
100& Q O Rctg.4 Scr Hammersley & {\bf{Meta Rctg}} & Rctg.4 Scr Halton & {\bf{Meta Rctg}} & Rctg.4 Scr Hammersley & O Rctg.7 Scr Halton & Rctg.4 Scr Halton & O Rctg.4 Scr Halton & Rctg.4 Scr Hammersley \\
\hline
150& O Rctg.4 Scr Hammersley & Q O Rctg.4 Scr Hammersley & {\bf{Meta Rctg}} & O Rctg.4 Scr Hammersley & Q O Rctg.7 Scr Hammersley & Rctg.4 Scr Halton & O Rctg.7 Scr Hammersley & {\bf{Meta Rctg}} & {\bf{Meta Rctg}} \\
\hline
600& Rctg.4 Scr Halton & Rctg.4 Scr Hammersley & O Rctg.4 Scr Hammersley & {\bf{Meta Rctg}} & {\bf{Meta Rctg}} & Rctg.4 Scr Halton & Rctg.4 Scr Hammersley & {\bf{Meta Rctg}} & Rctg.4 Scr Halton \\
\hline
\end{tabular}
    \caption{\label{ngoneshot} Artificial objective functions from Nevergrad (see text) with $P_0$ known: for each combination (dimension, budget), we mention the method which had the best frequency of outperforming other methods in that setting. 7400 replicas per run. O refers to opposite and QO refers to quasiopposite. We see that overall the Recentering (Rctg) methods perform best, with a parameter $k$ scaling roughly as Eq. \ref{metacalais} (we can see QO as further reducing the constant). Overall the Rctg reshaping outperforms its ancestor the  quasiopposite sampling~\cite{centerbased}. RctgX.Y refers to $\lambda=X.Y$. MetaRctg refers to \ot{$\lambda$ chosen by }Eq. \ref{metacalais}; it turns out to be one of the best methods overall, performing close to the best for each budget/dimension in the present context of $P_0$ known. These experiments use the artificial Nevergrad oneshot experiments, for which the distribution of the optimum is a standardized multivariate normal distribution. In this context, Cauchy distributions do not help much. {In bold the method that most often  performed best.}}
\end{table*}
\def\bennon{\begin{algorithm}[t]\small
\caption{\small $\LHS$ in dimension $1$ with sample sizes $2^k$ for $k>0$. This algorithm generates a sequence of points in dimension $1$, and each initial segment of length $2^k$ is a correct LHS sample. \label{simpleLHS}}\small
\begin{algorithmic}[1]
\Statex{{\bf Input:} $n=2^k$}
\If{$\textrm{list}$ does not exists}
\State{draw $r$ uniformly at random in $[0,1]$}
\State{$\textrm{list}\leftarrow[r]$}
\EndIf
\While{$n > length(\textrm{list})$}
\State{$\textrm{Supplementary\_list}\leftarrow[\ ]$ \Comment{empty list}}
\State{$\textrm{num\_intervals} \leftarrow 2\times length(\textrm{list})$}
\State{\Comment{multiply the length of $\textrm{list}$ by $2$}}
\For{$i\in \{0,\dots,\textrm{num\_intervals}-1\}$}
\State{$I=[i/\textrm{num\_intervals}, (i+1)/\textrm{num\_intervals}]$}
\If{$\textrm{list}$ has no point in $I$}
\State{add a uniform random point from $I$ in}
\State{~~~~$\textrm{Supplementary\_list}$}
\EndIf
\EndFor
\State{add $\textrm{Supplementary\_list}$ into $\textrm{list}$}
\EndWhile
\Statex{{\bf Return:} $\textrm{list}$}
\end{algorithmic}
\end{algorithm}}
\section{Introduction}\label{intro}
One-shot optimization, a critical component of hyperparameter search, consists in approximating the minimum of a function \ot{$f$ by its minimum $\min(f(x_1),\dots,f(x_n))$ over a finite subset $\{ x_1,\dots,x_n\}$ of} points provided by a sampler.
Space-filling designs such as Halton, Hammersley or Latin hypercube sampling, aim at distributing the points more diversely than independent random sampling. While their performance is well known for numerical integration~\cite{koksma}, their use for one-shot optimization is far less explored\cite{nie}. It was pointed out how much random sampling is sometimes hard to outperform~\cite{bergstra}. \cite{bousquet} advocated low discrepancy sequences, in particular Scrambled Hammersley\cite{hammersley,atanassov}. We quantify the benefit of such approaches \ot{and, looking for more headroom,} propose the concept of distribution reshaping, i.e. using a search distribution different from the prior distribution of the optimum.

{\bf{Contributions.}}
		\ot{Stochastic dispersion has been identified~\cite{bousquet} as a tool for measuring the performance of one-shot optimization methods.} We prove in Section \ref{sec:theory} 
		stochastic dispersion bounds of various sampling methods. These bounds are actually close to those of random search, including in the case of a limited number of critical variables~\cite{bousquet}, i.e. the case in which part of the variables have a strong impact on the objective functions whereas others have a negligible impact.
Given this limited \ot{headroom}, we propose reshaping (Section \ref{reshaping}), i.e. changing the search distribution, in two distinct flavors. First, even if the prior probability distribution of the optimum is known, \ot{we use a search distribution tightened }closer to the center as the dimension increases or as the budget decreases (Section \ref{calais}, recentering). Second \ot{(and possibly simultaneously, in spite of the apparent contradiction)}, we use Cauchy counterparts for searching closer to the boundaries (for bounded hyperparameters) or farther from the center (for unbounded hyperparameters). Experiments validate the approach (Section \ref{thexps}).
\def\cherrypixp{\begin{figure}[H]
    \centering
    \includegraphics[width=.41\textwidth, height=.26\textwidth]{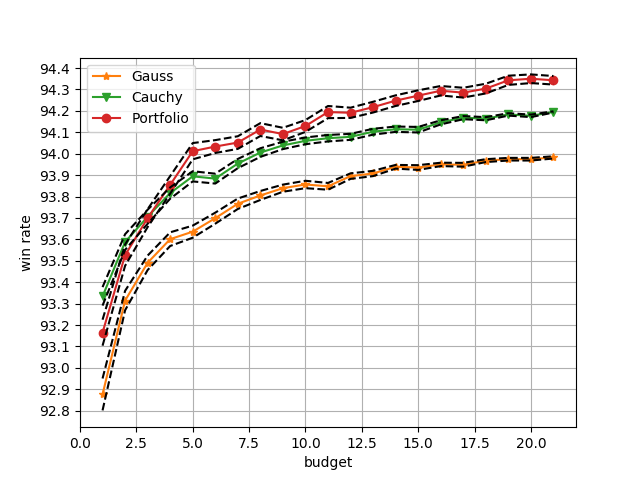}
    \caption{Cauchy vs Gauss vs Portfolio vs default parameters, for CherryPi, \ccc{ a bot playing Starcraft games}. Best random search value for various budgets, obtained by resampling. We observe better statistics for Cauchy and Portfolio. 
}
    \label{starcraft}
    \vspace{-2ex}
\end{figure}}
\def\removethat{{\bf Centers vs corners.} {We found} that sampling the center of the domain once is useful in high dimension. 
This is because, maybe counter-intuitively, each of $n$ (\omitme{standard} Gaussian) randomly drawn points will be far from the (\omitme{standard} Gaussian) optimum, way more than from $0$, if the dimension is large. 
We will also use the Rctg method, as an extension of this idea.}


{\bf{Space-filling vs reshaping: the two components of one-shot optimization.}}
\label{sec:sota}
Let us distinguish two probability distributions:
 the prior probability distribution $P_0$ of the optimum, and
 the search probability distribution $P_s$ used in the one-shot optimization method in particular in high dimension.
We show below that $P_s$ different \ot{from $P_0$ is theoretically and experimentally better, in particular in high dimension.}
While usual space-filling designs, compared to random search, relax the assumptions that samples are independently drawn according to $P_s$, we propose {(inspired by~\cite{centerbased,quasiopposite})} reshaped versions using $P_s$ \jr{spikier than $P_0$ around the center}.
 We provide in Table \ref{bigtable} an overview of theoretical results regarding space-filling designs without any reshaping component, i.e. $P_s=P_0$.
\def\prout{Besides all these theoretical results, we point out that  practitioners often
use Gaussian or uniform sampling. Actually, in many cases,
different distributions, aimed at perturbing a small number of variables, will perform  better. In fact, many variables should be set close to their default values, though we do not know which ones. }Regarding reshaping, maybe surprisingly, {\em{ even if the optimum is randomly drawn as a standard normal distribution}} in an artificial problem (e.g. we know a priori that the optimum $x^*$ is randomly drawn as a standard Gaussian and the objective function is $x\mapsto \|x-x^*\|^2$), the optimal search distribution {\em{ is not a Gaussian}} (Theorem \ref{midpoi}). This is the principle behind the addition of a middle point and, by extension, the principle of the Recentering reshaping (Section \ref{calais}). \ot{The Cauchy reshaping, on the other hand, is justified by distinct considerations, such as compensating the corner avoidance properties of some space-filling designs, and by the risk of expert errors on the correct range of an hyperparameter.}


\section{Space-filling designs}
 As recapped in Table \ref{bigtable},
we consider the following sampling strategies (space-filling designs), aimed at being ``more uniform'' than independent random search {e.g. in terms of dispersion, discrepancy, stochastic dispersion.} 
\paragraph*{Grid} picks up the middle of each of $k^d$ hypercubes covering the unit hypercube, with $k$ maximum such that $k^d\leq n$. We then sample $n-k^d$ additional random points uniformly in the domain. 

\omitme{\textbf{Pure random search} is the standard baseline.} 

\paragraph*{Latin Hypercube Sampling (LHS)}~\cite{eglajs,mckay} defines $\sigma_1,\dots,\sigma_d$, random independent permutations of $\{0,\dots,n\}$ and then for $0\leq i \leq n$, the $j^{th}$ coordinate of the $i^{th}$ point of the sequence is given by ${(x_i)_j=(\sigma_j(i) + r_{i,j})/(n+1)}$ where the $r_{i,j}$ are independent identically distributed, uniformly in $[0,1]$. 

\paragraph*{Jittered sampling} consists in splitting $\mathcal{D}$ into $n=k^d$ (assuming that such a $k$ exists) hypercubes of volume $1/k^d$ and drawing one random point uniformly in each of these hypercubes~\cite{pausinger}. Other forms of jittered sampling exist, e.g., with different number of points per axis. \omitme{The generalization of the following results to such a case are straightforward.}
 \paragraph*{Quasi-random} (QR) are also called \textit{low discrepancy sequences}, due to some good distribution properties in the domain\omitme{(Section \ref{sec:discrepancy})}. This is the case for \textit{Halton}, \textit{Hammersley} and \textit{Sobol} sequences. {\textbf{Halton}~\cite{halton60} defines the $j^{th}$ coordinate of the $i^{th}$ point of the sequence as ${(x_i)_j=radixInverse(i, p_j)}$ where}:
(1) $p_0,\dots,p_{d-1}$ are coprime numbers. Typically, but not necessarily, $p_i$ is the $(i+1)^{th}$ prime number.
(2) ${radixInverse(k, p)=\sum_{j\geq 0} a_j p^{-j}}$ with $(a_j)_{j\geq 0}$ being the writing of $k$ in basis $p$, i.e., ${k=\sum_{j\geq0} a_jp^j}$.
\textbf{Hammersley's}  sequence is given by ${(x_i)_j=radixInverse(i, p_{j-1})}$ when $j>0$ and ${(x_i)_0=\frac{i+\frac12}n}$, see~\cite{hammersley}.
\textbf{Sobol}'s sequence~\cite{sobol67} is another advanced quasi-random sequence.
\paragraph*{Modifiers.}
We use various modifiers of our samplers: scrambling~\cite{atanassov,permut1,owen}, applied to our Halton and Hammersley implementations, and generic modifications applicable to all samplers: random shift (adding a random vector in the unit hypercube and applying modulo 1)~\cite{tuffinvariance}. 
Other modifications are based on reshaping (Section \ref{reshaping}).

\paragraph*{Performance measure and proxies.} We use the classical notions of simple regret, discrepancy (with respect to axis-parallel rectangles and the $L^\infty$ norm), dispersion and low-discrepancy sequences (i.e. sequences with discrepancy decreasing as $O(\log(n)^d/n)$).
We refer to \cite{nie} or the supplementary material (SM) for classical notions such as discrepancy and dispersion, and to \cite{bubeck2009pure} for an overview of the literature on simple regret.
Discrepancy is well known as a good criterion for numerical integration~\cite{koksma}. It was recently proposed as a criterion also for one-shot optimization~\cite{bergstra,bousquet}.  
Given a domain $\D$, the stochastic dispersion~\cite{bousquet} of a random variable ${X=(x_1,\dots,x_n)}$ in $\D^n$ is defined as 
$\sdisp(X)=\sup_{x^*\in D} \E_{X} \inf_{1\leq i\leq n} \| x_i - x^*\|$.

 \paragraph*{Critical vs useless variables.} We distinguish cases in which there are $d'$ unknown variables (termed critical variables) with an impact on the objective function whereas $d-d'$ variables have no impact. 

\def\noincrementality{
{\bf{Incrementality}} is another feature of samplers. It refers to the capability of a sampler to extend a previously generated sample of size $n$ to a sample of size $n+k$ without performance loss compared to the direct generation of $n+k$ points. {The point in one-shot optimization (as opposed to e.g. Bayesian optimization) is precisely that proposed hyperparameters can be evaluated all in parallel; however,} incrementality matters in case $k$ free workers suddenly come up. {A sampler which maps $n$ to a random sample $X_n\in \D^n$ is said to be $n+k$ incremental if, for all $n$, there exists a mapping $m_{m,k}$ from $\D^n\to \D^{n+k}$ such as the probability distribution of $m_{m,k}(X_n)$ has the same probability distribution as $X_{n+k}$.}}

\section{Theory: are space-filling designs all equal ?}
\label{sec:theory}
\ot{Table \ref{bigtable} shows} bounds on the performance of various space-filling designs, including the very classical LHS, the top performing (in terms of discrepancy) Scrambled-Hammersley, and \jr{the robust} jittered sampling. We show that they perform well but just up to a constant factor, compared to random search, even in the case of critical variables: this is the key reason for introducing reshaping in Section \ref{reshaping}.
For simplicity, we always consider $\mathcal{D}=[0,1]^d$ as the search domain and $n$ the number of points in the sequence. 
{See SM for other results and detailed proofs.}

\subsection{Latin hypercube sampling (LHS): stochastic dispersion}\label{sec:LHS}

{LHS{~\cite{mckay,eglajs}} is particularly appreciated for building a surrogate model, e.g. in Efficient Global Optimization~\cite{jones}. Due to its popularity, some variants have been developed to generate better LHS, based for example on the maximin distance criterion~\cite{johnson}, or for sliced LHS~\cite{qian}. 
}
\cite{doerr,lhsr} have established discrepancy properties of LHS, in particular for moderate budgets.
We show that the stochastic dispersion of LHS is optimal, i.e., it decreases at the optimal rate $O(1/n^{1/d})$ {(see Prop.~\ref{zeporp})}. \ot{Importantly, LHS is entirely preserved by projection to any subspace - as opposed to low discrepancy sequences, for which only segments of initial variables keep the same constants in discrepancy or dispersion bounds.} 
\def\tosm{
\begin{prop}\label{prop:LHS2n}
LHS is $2n$-incremental.
\end{prop}
The \omitme{detailed} proof is given in SM. 
Considering the non-zero probability of having two points in $1/(n+k),2/(n+k)${, we show} that it is not possible to get a $n+k$-incrementality.
\begin{prop}\label{prop:LHS2}
{For all $k>0$, $\LHS$ is not $(n+k)$-incremental.} {Also for all $k>0$} it is not $(2n-k)$-incremental and it is not $\lfloor \alpha n \rfloor$-incremental for $1<\alpha<2$.
\end{prop}

}
\begin{prop}\label{zeporp}
Consider $m\leq n$ two powers of $2$ and ${\alpha=m/n^{1-1/d}}$.
Then, the probability $P_{n,m}$ that $\LHS(n)$ does not intersect $[0,m/n]^d$ is at most $\left(1-\frac{\alpha^{d-1}}{n^{1-1/d}}\right)^{\alpha n^{1-1/d}}$.
\end{prop}

As a by-product of this proposition, the SM shows (limited) incrementality properties of LHS.
Property \ref{zeporp} has implications in terms of distance to an arbitrary $x$ in $[0,1]$.
The probability that $x$ has no neighbor at distance (for the maximum norm) $m/n$ is at least the probability that $\LHS(n)$ has no point in the hypercube $[0,m/n]^d$.

$P_{n,m}$ is $0$ if $d=1$,
and, if $d>1$ and $m=\alpha n^{1-1/d}$, then $P_{n,m}$ converges to $\exp(-\alpha^d)$.
{In other words, the quantile $1-\delta$ of the minimum distance to $x$ is $O(\sqrt{d}\log(1/\delta)^{1/d}/n^{1/d})$ - the same as random search, up to the constant (next Section).}  


\ifthenelse{\incrementality=1}{
\def\prout{
\subsection{Strong non-incrementality of Hammersley}\label{sec:hammersley}

We extend known conjectures (actually conjectures or theorems, depending on the dimension) on the non-incrementality of Hammersley's sequence to more general forms of incrementality.\omitme{ (i.e.} {Specifically,} we show that the known conjectures imply non-incrementality beyond the classical non-$(n+1)$-incrementality.
Conjectures on the lower bound of the discrepancy are as follows.

\begin{conj}\label{conjcd}
For any finite point set ${x_1,x_2,\dots,x_n}$, there exists a constant $c_d$ depending only on the dimension $d$ such that $\disc(x_1,x_2,\dots,x_n) \geq c_d\frac{(\log n)^{d-1}}{n}.$
\end{conj}

\begin{conj}\label{conj:2}
For any dimension $d$, there exists a constant $c'_d>0$ depending only on $d$ such that for any infinite sequence ${x_1,x_2,\dots}$, there exist infinitely many $n$ such that 
$\disc(x_1,x_2,\dots,x_n) \geq c'_d\frac{(\log n)^{d}}{n}.$
\end{conj}

These conjectures~\cite{conja} are equivalent and have been proved~\cite{sch} for $d=2$. They are often found in the literature for the star-discrepancy, but also hold for the discrepancy as the discrepancy and star-discrepancy are equivalent up to a factor $2^d$. \nico{citation needed}

{
\begin{prop}[Assuming Conjecture \ref{conj:2} in dimension $d$]\label{prop:hammersley}
Consider a dimension $d$, $k\in \{1,\dots,n-1\}$ and $0<\epsilon\leq 1$.
The Hammersley point sets in dimension $d$ are not $\left({n + k \lfloor log(n)^{d-\epsilon} \rfloor}\right)$-incremental.
\end{prop}}


\begin{remark}
More generally, Conjecture \ref{conj:2} implies that any sequence $X_n$ of $n$ points in dimension $d$ satisfying
$disc(X_n)=O\left(\frac{(\log n)^{d-1}}{n}\right)$
is not $\left({n + k \lfloor log(n)^{d-\epsilon} \rfloor}\right)$-incremental, for all $k\in \{1,\dots,n-1\}$ and $0<\epsilon\leq 1$.
\end{remark}%
}
}{}

{\bf{Dispersion of projected Jittered Sampling}}\label{sec:jittered}
In this section, we show that Jittered Sampling benefits from the same stochastic dispersion bounds as random search.
We use $V_d$ the volume of $\{x\in \R^d; \|x\|\leq 1\}$.
We first mention a known property of random search (details in next section), useful for proving Prop. \ref{jitproj}.
\begin{lemma}[Random search]\label{rs}
Consider $x$ an arbitrary point in $[0,1]^{d}$.
Consider $x_1,\dots,x_M$ a sequence generated by random search.
Define ${\e\omitme{:=\e_{rs}}=\min_i \|x_i-x\|}$. Then,
$$\forall \delta, P\left(\e>2^{1+1/d} \left(\log(1/\delta)^{1/d}/(MV_{d})^{1/d} \right)\right)\leq \delta.$$
\end{lemma}

\begin{prop}[Jittered sampling has optimal dispersion after projection on a subset of axes]\label{jitproj}
Consider $x\in[0,1]^{d'}$.
Consider jittered sampling with $n=k^d$ points. Consider its projection on the $d'<d$ first coordinates\footnote{Without loss of generality; we might {consider a projection to $d'$ arbitrarily} chosen coordinates.}. Let $x_1,\dots,x_n$ be these projected points.
Define $\e=\min_i \|x_i-x\|$. Then,
$\mbox{with probability at least $1-\delta$,}\e\leq 2^{1+1/d'} \frac{\log(1/\delta)^{1/d'}}{(V_{d'}n)^{1/d'}}.$
\end{prop}
The quantile $1-\delta$ of this minimum distance, up to constant factors, is therefore the same as for a LHS sample of cardinal $n$ or for a pure random sample, namely $O(\sqrt{d}\frac{\log(1/\delta)^{1/d'}}{n^{1/d'}})$.
\def\removeoldgans{
\begin{table}[t]
\tiny\center
\begin{tabular}{|c|c|c|c|}
\hline
Average min  & Average min	& Average min  &GAN Model\\
 FID LHS & FID QR	& FID R	&  \\	
\hline		
\multicolumn{4}{|c|}{First version of the platform}\\
\hline
{\bf{33.56}}	&	36.73	&	39.37		&	BEGAN		\\
{\bf{23.72}}	&	27.22	&	26.32		&	DRAGAN		\\
{\bf{23.92}}	&	25.44	&	25.07	&GAN		\\
26.30	&	{\bf{25.81}}	&	29.09		&	GAN-MINMAX		\\
{\bf{25.00}}	&	25.40	&	25.04			& LSGAN			\\
56.77	&	{\bf{56.45}}	&	56.60	&	VAE		\\
23.62	&	{\bf{21.63}}	&	26.81	&	WGAN				\\
{\bf{21.66}}	&	22.48	&	23.42	&WGAN-GP			\\
\hline
\multicolumn{4}{|c|}{Second version of the platform, where GAN models were improved}\\
\hline
 Average min  & Average min 	& Average min   & GAN Model\\
 FID LHS & FID QR	& FID R	&  \\	
\hline										
{\bf{21.12}}	&	35.91	&	36.97	&		BEGAN				\\
25.10	&	{\bf{24.56}}	&	29.18	&		DRAGAN			\\
28.236	&	{\bf{24.14}}	&	26.35	&	GAN				\\
{\bf{27.29}}	&	27.91	&	28.97	&		GAN-MINMAX				\\
	26.78	&	25.81	&	{\bf{25.64}}	& LSGAN				\\
55.80	&	55.81	&	{\bf{55.71}}	& VAE				\\
20.96	&	{\bf{17.32}}	&	19.22	&	WGAN				\\
{\bf{20.51}}	&	21.64&	26.07	&		WGAN-GP				\\
\hline
\end{tabular}
	\caption{\label{resganLHS}{Comparison of LHS vs QR vs R in terms of Frechet Inception Distance (FID) for various GAN models trained on Fashion MNIST, using 100 samples. The two subtables correspond to two versions of the platform: GANs models were improved in the mean time. Overall, for a budget of 100 samples, LHS outperformed random search in 11 out of 16 rows and QR outperformed random in 11 out of 16 rows; this is consistent with the usual $\simeq\frac23$ corresponding to  space-filling designs reaching the same performance with twice less resources than random search. \omitme{- random search with budget $2n$ outperforms random search with budget $n$ with probability exactly $\frac23$}} 
 \label{resganqr}  }
\end{table}}



\section{Theory of reshaping: middle point \& recentering}\label{addmiddle}\label{reshaping}
Interestingly, in the context of the initialization of {differential evolution~\cite{de}}, \cite{centerbased} proposed a sampling focusing on the middle. Their method consists in adding for each point $x$ in the domain $[-1,1]^d$, a point $-r\times x$ for $r$ uniformly drawn in $[0,1]$. This combines {\bf{opposite sampling}} (which corresponds to antithetic variables, also used for population-based optimization in \cite{antithetic}) and focus to the center (multiplication by a constant $<1$): their method is termed {\bf{quasiopposite}} sampling.
Consider dimension $d$ and $x^*$ randomly normally distributed with unit variance in $\R^d$.
Consider $x_1,\dots,x_n$, independently randomly normally distributed with unit variance in $\R^d$.
The median of $\|x^*\|^2$ is denoted $m_0$ and the median of $\min_{i\leq n} \| x_i-x^* \|^2$ is denoted $m_n$. We first note lemmas \ref{mlm1} and \ref{mlm2}.
\begin{lemma}\label{mlm1}
$m_0$ is equivalent to $d$ as $d\to \infty$.
\end{lemma}
\begin{lemma}\label{mlm2}
{$P(\|x_1-x^*\|^2\leq m_n)\geq \frac{1}{2n}$.}
\end{lemma}
{We now note that $\left\|\frac{1}{\sqrt{2}}(x_1-x^*)\right\|^2$ follows a $\chi^2$ distribution with $d$ degrees of freedom.}
\begin{lemma}\label{lem:chernoff}
{By Chernoff's bound for the $\chi^2$ distribution,
$P\left( \|x_1-x^*\|^2 \leq d(1+o(1)) \right)\leq \left((1+o(1))\frac12\exp(\frac12)\right)^{d/2}$.}
\end{lemma}

\begin{theorem}\label{midpoi}
Consider $n>0$. There exists $d_0$ such that for all $d>d_0$,
if $X$ is a random sample of $n$ independent standard normal points in dimension $d$,
if $x^*$ is a random independent normal point in dimension $d$,
then, unless ${n\geq \frac12 \left((1+o(1))\frac12\exp(\frac12)\right)^{-d/2}}$, the median of the minimum distance $\min_{x\in X} \|x-x^*\|$ is greater than the median of the distance $\|x^*-(0,\dots,0)\|$.
\end{theorem}

{As having $n$ points equal to $0$ is pointless, we can consider $n-1$ standard independent normal points, plus a single middle point: this is our modification ``plus middle point''. We will propose another related reshaping, namely tightening the distribution closer to the center: the Recentering method. Then we will see a distinct reshaping, namely switching to the Cauchy distribution.}


\def\omitthis{\begin{remark}
\cite{elements} states that in high dimension, uniform random search in a unit ball samples points closer to the boundary than to the center. Formally, the median distance from the origin to the closest data point is given by $\left(1-\frac{1}{2}^{1/n}\right)^{1/d}$
where $n$ is the sample's size and $d$ is the dimension. 
\end{remark}}

 

\label{calais}

\begin{table*}[t]
    \centering\scriptsize
\begin{tabular}{|P{0.03\textwidth}|P{0.087\textwidth}|P{0.087\textwidth}|P{0.087\textwidth}|P{0.087\textwidth}|P{0.087\textwidth}|P{0.087\textwidth}|P{0.087\textwidth}|P{0.087\textwidth}|P{0.087\textwidth}|P{0.087\textwidth}|}
\hline
 & 25&50&100&200&400&800&1600&3200&6400&12800\\
\hline
10& Hmsley Plus Middle Point & Scr Hmsley Plus Middle Point & Scr Hmsley & Rctg.7 Scr Halton & Halton & Meta Rctg & Random Plus Middle Point & L H S & Halton & Halton Plus Middle Point \\
\hline
20& Scr Hmsley & Scr Halton Plus Middle Point & Scr Halton & Scr Halton & Halton Plus Middle Point & L H S & O Random & L H S & Hmsley & Hmsley \\
\hline
30& Rescale Scr Hmsley & {\bf{Cchy Rctg.55 Scr Hmsley}} & {\bf{Cchy Rctg.55 Scr Hmsley}} & {\bf{Cchy Rctg.55 Scr Hmsley}} & {\bf{Cchy Rctg.55 Scr Hmsley}} & Meta Cchy Rctg & {\bf{Cchy Rctg.55 Scr Hmsley}} & {\bf{Cchy Rctg.55 Scr Hmsley}} & {\bf{Cchy Rctg.55 Scr Hmsley}} & Cchy Rctg.4 Scr Hmsley \\
\hline
40& Scr Hmsley & Scr Hmsley Plus Middle Point & Scr Halton Plus Middle Point & Scr Hmsley Plus Middle Point & Scr Halton Plus Middle Point & Random & Scr Hmsley & L H S & Random Plus Middle Point & Scr Hmsley Plus Middle Point \\
\hline
60& Rescale Scr Hmsley & {\bf{Cchy Rctg.55 Scr Hmsley}} & Cchy Rctg.7 Scr Hmsley & {\bf{Cchy Rctg.55 Scr Hmsley}} & Cchy Rctg.7 Scr Hmsley & {\bf{Cchy Rctg.55 Scr Hmsley}} & {\bf{Cchy Rctg.55 Scr Hmsley}} & Meta Cchy Rctg & {\bf{Cchy Rctg.55 Scr Hmsley}} & Cchy Rctg.4 Scr Hmsley \\
\hline
120& Rescale Scr Hmsley & Cchy Rctg.7 Scr Hmsley & Cchy Rctg.7 Scr Hmsley & Cchy Rctg.7 Scr Hmsley & {\bf{Cchy Rctg.55 Scr Hmsley}} & {\bf{Cchy Rctg.55 Scr Hmsley}} & {\bf{Cchy Rctg.55 Scr Hmsley}} & {\bf{Cchy Rctg.55 Scr Hmsley}} & {\bf{Cchy Rctg.55 Scr Hmsley}} & {\bf{Cchy Rctg.55 Scr Hmsley}} \\
\hline
\end{tabular}    
    \caption{\label{rwoneshot} Experiments on the Nevergrad real-world rescaled testbed. This experiment termed ``oneshotscaledrealworld'' corresponds to real-world test cases in which a reasonable effort for rescaling problems according to human expertise has been done; $P_0$ is not known, but an effort has been made for rescaling problems as far as easily possible. Dimension from 10 to 120, budget from 25 to 12800. There is no prior knowledge on the position of the optimum in this setting; MetaRctg did not perform bad overall but Cauchy variants dominate many cases, as well as rescaled versions which sample close to boundaries. Hmsley stands for Hammersley, Cchy for Cauchy, Rctg for Rctg. In bold the method performing best overall; MetaCauchyRctg performs well overall though other CauchyRctg often performed better - most often with a constant $<1$, i.e. recentering, and most often with Cauchy (all samplers are tested in both flavors, normal and Cauchy) for high dimension, which validates both Recentering and Cauchy.}
\end{table*}
\begin{figure}[t]
    \centering
    \includegraphics[width=.45\textwidth,height=.22\textwidth,trim={3 3 3 3},clip]{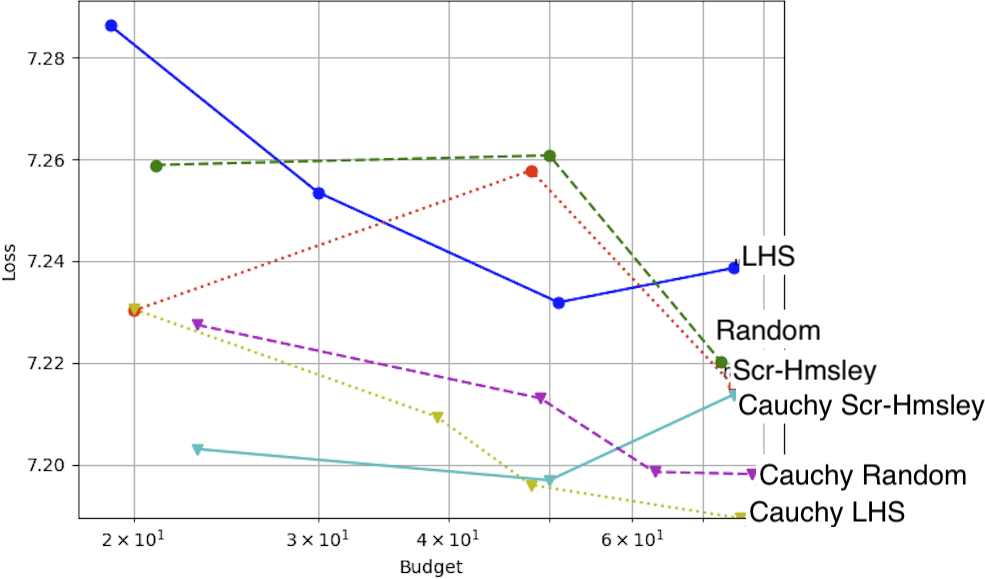}\\
    \caption{\label{air12}Cauchy vs Gauss on the Attend/Infer/Repeat model. For each setting in \{random, LHS, ScrambledHammersley\}, we reach better loss values when switching to the Cauchy counterpart. \ot{ P-value 0.05 for Cauchy vs Normal.}  See SM for validation in terms of counting objects.
    }
\end{figure}
\begin{table*}[t]\centering
\scriptsize
\begin{tabular}{|P{0.025\textwidth}|P{0.091\textwidth}|P{0.091\textwidth}|P{0.091\textwidth}|P{0.091\textwidth}|P{0.091\textwidth}|P{0.091\textwidth}|P{0.091\textwidth}|P{0.091\textwidth}|P{0.091\textwidth}|P{0.091\textwidth}|}
\hline
 & 25&50&100&200&400&800&1600&3200&6400&12800\\
\hline
10& Scr Halton Plus Middle Point & Rescale Scr Hmsly & Scr Hmsly & Rescale Scr Hmsly & Meta Ctrg & Rescale Scr Hmsly & Rescale Scr Hmsly & Scr Hmsly & L H S & Scr Hmsly Plus Middle Point \\
\hline
15& Cchy Scr Hmsly & Cchy Random & Cchy LHS & Cchy LHS & Ctrg20 Scr Halton & Cchy Scr Hmsly & Cchy Scr Hmsly & Cchy Ctrg12 Scr Hmsly & Cchy Scr Hmsly & Cchy Ctrg12 Scr Hmsly \\
\hline
20& L H S & Random & Rescale Scr Hmsly & Rescale Scr Hmsly & Rescale Scr Hmsly & Rescale Scr Hmsly & Rescale Scr Hmsly & Rescale Scr Hmsly & Rescale Scr Hmsly & Hmsly Plus Middle Point \\
\hline
30& Cchy Random & Cchy LHS & Cchy LHS & Cchy Random & Cchy Random & Cchy LHS & Cchy Random & Cchy LHS & Cchy Random & Cchy Random \\
\hline
40& Ctrg12 Scr Halton & Rescale Scr Hmsly & Rescale Scr Hmsly & Rescale Scr Hmsly & Rescale Scr Hmsly & Rescale Scr Hmsly & Rescale Scr Hmsly & O Random & Rescale Scr Hmsly & Random \\
\hline
60& Cchy LHS & Cchy Random & Cchy LHS & Cchy Ctrg7 Scr Hmsly & Cchy LHS & Cchy Ctrg7 Scr Hmsly & Cchy Random & Cchy LHS & Cchy LHS & Cchy Random \\
\hline
120& Cchy LHS & Cchy LHS & Cchy LHS & Cchy Random & Cchy LHS & Cchy Random & Cchy Random & Cchy Random & Cchy Random & Cchy LHS \\
\hline
675& Cchy Ctrg12 Scr Hmsly & Cchy Ctrg4 Scr Hmsly & Cchy Scr Hmsly & Cchy Random & Cchy Ctrg7 Scr Hmsly & Cchy Ctrg4 Scr Hmsly & Cchy Ctrg4 Scr Hmsly & O Ctrg20 Scr Halton & Cchy Scr Hmsly & Meta Cchy Ctrg \\
\hline
\end{tabular}
\caption{Counterpart of Table \ref{rwoneshot} with the experiment termed ``oneshotscaledrealworld'' in Nevergrad, which contains more problems including many for which no rescaling effort has been made. Cntrg does not make sense in this ``totally unknown $P_0$'' setting, but we still see a lot of Cauchy counterparts or Rescaled versions, showing that focusing close to boundaries (for bounded hyperparameters) or on large values (for unbounded hyperparameters) makes sense. \ot{The contrast with Table \ref{rwoneshot} in which CauchyMetaCtrg and close variants such as CauchyCtrg with $\lambda=.55$ perform well suggest that scaling parameters and data is a good idea for generic methods to be effective.} \label{unscaled}}
\end{table*}
{\bf{Recentering (Rctg for short) reshaping.}} Consider optimization in $[0,1]^d$. Given a sample $s$, and using $g$ the cumulative distribution function of the standard Gaussian, the Rctg reshaping consists in concentrating the distribution towards $(0.5,\dots,0.5)$: it considers $\{c(s) \mid s\in S\}$ rather than $S$, with $c(s)=g(\lambda\times g^{-1}(s_1)),\dots,g(\lambda\times g^{-1}(s_d))$. $\lambda=0$ sets all points to the middle of the unit hypercube. $\lambda=1$ means no reshaping. 
{\bf{MetaRecentering.}} Preliminary experimental results lead to the specification of MetaRctg, using the dimension and the budget for choosing the parameter $k$ of the Recentering reshaping:
MetaRctg uses Scrambled-Hammersley and Rctg reshaping with 
\begin{equation}
\lambda=\frac{1+\log(budget)}{4\log(dimension)}.\label{metacalais}
\end{equation}
{\bf{Cauchy.}} When using the Cauchy distribution, we get  $c(s)=g(\lambda\times C^{-1}(s_1)),\dots,g(\lambda\times C^{-1}(s_d))$ with $C$ the Cauchy cumulative distribution function. 
{\bf{Extension to unbounded hyperparameters.}} $g(.)$ can be removed from those equations when we consider sampling in $\R^d$ rather than in $[0,1]^d${: we then get $ c(s)=\lambda\times g^{-1}(s_1),\dots,\lambda\times g^{-1}(s_d)$ in the normal case.} $g$ can also be applied selectively on some variables and not on others when we have both bounded and unbounded hyperparameters.

\section{Experiments}\label{thexps}
{See SM for reproduction of all our artificial experiments with one-liners in the Nevergrad platform~\cite{nevergrad}, and for additional deep learning experiments with \F2F5~\cite{f2f5}.}
We first test our baselines in an artificial setting with $P_0$ known and without any reshaping (Section \ref{sec:experiments}). We then check that Recentering works in such a context of $P_0$ known (Section \ref{rctgsec}).
Then we switch to $P_0$ unknown (Section \ref{nopo}). We first check the impact of Cauchy (Section \ref{sec2}). We then check Recentering and Cauchy simultaneously (Section \ref{sec1} and \ref{sec3}). We conclude that Cauchy is vastly validated for unknown $P_0$ and that Recentering works if $P_0$ is known {\em{or if underlying data are properly rescaled (Table \ref{rwoneshot}, as opposed to the wildly unscaled problems in Table \ref{unscaled})}}.

\subsection{When the prior $P_0$ is known}
There are real world cases in which $P_0$ is known;
typically when optimizations are repeated: maximum likelihood in item response theory repeated for estimating the parameters of many questions, ELO evaluation of many gamers from their records, hyperparametrization of cloud-based machine learning~\cite{cloudml}, repeated optimization of industrial oven parameters~\cite{indoven2} for distinct scope statements. \ot{Another important case is when the objective function is the worst outcome over a family of scenarios, to be approximated by a finite sample corresponding to ranges of independent exogeneous variables (dozens of annual weather parameters and financial parameters), as usually done in network expansion planning~\cite{scenar,scenar2}.}

\subsubsection{One-shot numerical optimization of artificial test functions with $P_0$ known and $P_s=P_0$: the baselines}\label{sec:experiments}
{{While we use mainly real-world experiments in the present paper, we draw the following conclusions from synthetic experiments with controlled $P_0$ and without reshaping, using classical objective functions from the derivative-free literature, various budgets and all samplers defined above. Detailed setup and results reported in SM; they confirm ~\cite{bergstra,bousquet} as follows:}  
\begin{itemize}
\item {Many low discrepancy methods (e.g. Halton, Hammersley and their scrambled counterparts) depend on the order of hyperparameters - intuitively they are ``more'' low discrepancy for the first variables.} Our experiments in optimization confirm that low discrepancy methods are better when variables with greater impact on the objective function are first. With scrambled low-discrepancy methods, results are less penalized (but still penalized) when important variables are last. 
\item {A strength of LHS is its independence to the order of variables and (as a consequence) its strong performance for a small number of randomly positioned critical variables.}
\end{itemize} {Besides confirming the state of the art, these experiments also confirm that adding a single middle point helps, in particular in high dimension. 
Among methods using $P_s=P_0$, Scrambled-Hammersley plus middle point is one of the best methods in this simplified setting.}
 \omitme{As presented in the SM, \cite{nevergrad} presents averaged results over a wide range of testbeds, when there is no useless variable. Scrambled Hammersley plus middle point - precisely the method advocated in~\cite{bousquet} and theoretically optimal from a point of view of discrepancy, plus our middle point modification. We also see in experiments with useless variables how much Cauchy and Portfolio samplings perform well. 
} 
}
{

\def\bennonvoila{\subsubsection{Initialization of Bayesian optimization: space-filling outperforms random search and adding a middle point helps}
\label{boinit}
Population-based optimization is significantly improved by a better initial population~\cite{feurer2015initializing,inoculation2,centerbased,qrinit}.
\cite{centerbased} has in particular shown that for the differential evolution algorithm~\cite{de}, focusing on the center by quasiopposite sampling is effective. We show in the SM that space-filling designs also work for initializing Bayesian optimization (already known), and that adding a middle point is also effective.}
}
 {
\subsubsection{Experiments including a reshaped space-filling design: Recentering works}\label{rctgsec}
\ot{We present an experiment on objective functions Sphere, Rastrigin and Cigar, with {100\% or 16.67\% of critical variables (i.e. in the latter case we add 5 randomly positioned useless variables for each critical variable), with budget in 30, 100, 300, 1000, 3000, 10000, 30000, 100000, 300000, with 3, 25 or 100 critical variables}.
We compare Rctg reshaping with constant $k\in \{0.01,0.1,0.4,0.55,1.0,1.2,2.0\}$, the same Rctg reshaping plus opposite sampling or quasi-opposite sampling, on top of LHS, Scrambled-Halton, Scrambled-Hammersley or pure random sampling; we consider Gauss or Cauchy as conversions to $\R^d$. Given the choice regarding critical variables previously mentioned, we get dimension 3, 18, 25, 100, 150, 600.
This is reproducible with a one-liner ``oneshotcalais'' in Nevergrad (see SM).}
\def\bobennon{  This extends the ``oneshot'' experiment originally in Nevergrad~\cite{nevergrad} by adding Rctg optimizers as in Section \ref{calais}. 
}
Results are presented in Table \ref{ngoneshot} and validate Recentering, in particular in its Meta version (Eq. \ref{metacalais}), as soon as $P_0$ is known. They also show that in such a context ($P_0$ perfectly known) Cauchy is not useful.
\subsection{Cauchy-reshaping with $P_0$ unknown}\label{nopo}
\subsubsection{No $P_0$: Cauchy for attend/infer/repeat} \label{cauchy}\label{sec2}
Previous results have validated Recentering (our first proposed reshaping) in the case of $P_0$ known. The second form of reshaping consists in using the Cauchy distribution when $P_0$ is unknown: we here validate it. As detailed in Section \ref{reshaping}, Cauchy makes sense also without recentering and for bounded hyperparameters (in that case, it increases the density close to the boundaries).
Fig.~\ref{air12} presents results on the Attend, Infer, Repeat (AIR) image generation testbed~\cite{AIR}.
AIR is a special kind of variational autoencoder, which uses a recurrent inference network to infer both the number of objects and a latent representation for each object in the scene. 
We use a variant that additionally models background and occlusions\ccc{, where details are provided in SM.}
We have 12 parameters, 
initially tuned by a human expert, namely 
the learning rates of the main network and of the baseline network, 
the value used for gradient clipping, 
the number of feature maps, 
the dimension of each latent representation, 
the variance of the likelihood,  
the variance of the prior distribution on the scale of objects, and 
the initial and final probability parameter of the prior distribution on the number of objects present. 
The loss function is the Variational Lower Bound, expressed in bits per dimension. 
{The dataset 
consists in 50000 images from Cifar10~\cite{cifar10} 
and 50000 object-free patches from COCO~\cite{coco}, split into balanced training (80\% of the samples) and validation sets.}
For each space-filling method, the Cauchy counterpart outperforms the original one. 
\subsubsection{Generative adversarial networks (GANs).}\label{gan}\label{pgz}\label{sec1}
We use Pytorch GAN Zoo~\cite{pytorchganzoo} for progressive GANs~\cite{pgan}, with a short 10 minutes training on a single GPU.
We optimize 3 continuous hyperparameters, namely leakiness of Relu units in $[1e-2,0.6]$, the discriminator $\epsilon$ parameter in $[1e-5,1e-1]$, and the base learning rate in $[1e-5,1e-1]$. MetaCauchyRctg performed best (Fig. \ref{pgzfig}), in spite of the fact that it was designed just by adding Cauchy to MetaRctg which was designed on independent distinct artificial experiments in {Tab. \ref{ngoneshot}}.
\begin{figure*}[t]
    \centering
    \includegraphics[width=.7\textwidth,trim={0 0 0 0},clip]{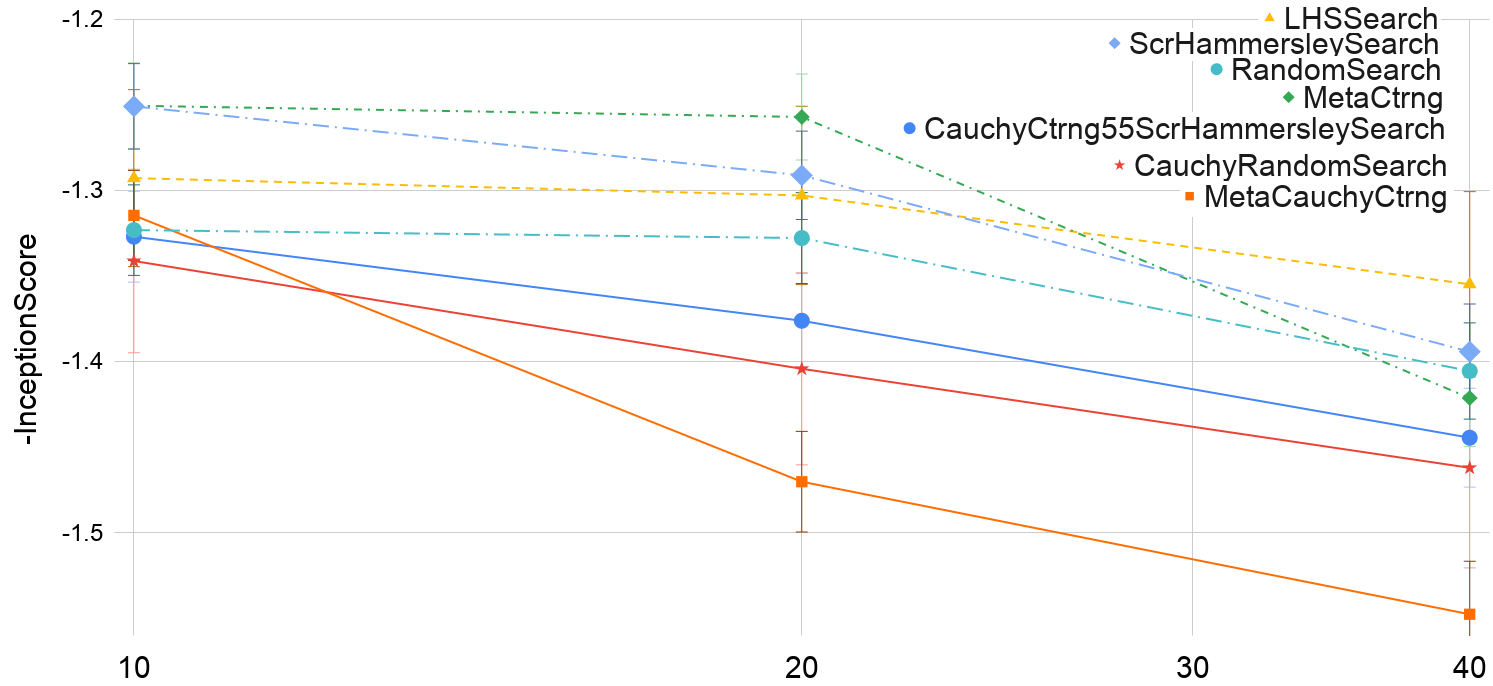}
    \caption{Experiments on progressive GANs\cite{pgan,pytorchganzoo}; see a discussion of criteria in \cite{borji2018pros}. MetaCauchyRctg dominates. All Cauchy variants (full lines) perform well compared to normal ones (dashed). The ``Meta'' choice of the recentering constant (Eq. \ref{metacalais})  seemingly performs well though without Cauchy and low budgets this was not that clear. X-axis = budget. Y-axis = opposite of inception score, the lower the better (we use opposite for consistency with other plots in the present paper, which consider losses to be minimized).}
    \label{pgzfig}
\end{figure*}
\subsubsection{Nevergrad real world experiments without control of $P_0$}\label{ngoneshotsec}\label{sec3}
Nevergrad provides a real world family of experiments, built on top of the MLDA testbed~\cite{mlda}\ot{, using Salmon mappings and clustering and others}: these experiments are less expensive and fully reproducible (see SM). We run all our one-shot optimizers on it with 7400 repetitions. We then check for each budget/dimension which method performed best (Tab. \ref{rwoneshot}): in this real-world context, in which assuming a standard deviation of $1$ for the uncertainty over the optimum is risky, the Cauchy sampling with Rctg reshaping 0.55 performs quite well. It is sometimes outperformed by rescaled Scrambled-Hammersley, which takes care of pushing points to the frontier (each variable is rescaled so that the min and the max of each variable, over the sample, hit the boundaries). This means that the best methods frequently have samplings able to work far from the center (either heavy tail for Cauchy, or sample {rescaled} for matching the boundaries). The exact optimal constant $\lambda$ does not always match the MetaRecentering scaling we have proposed (Eq. \ref{metacalais}) but is clearly $<1$. Though this setting already does not really have a known $P_0$ (just a rough rescaling of underlying data), we switch to a more hardcore setting in Table \ref{unscaled}: we consider the ``oneshotunscaledrealworld'' experiment in Nevergrad for a case in which no rescaling effort was made; we still get a quite good behavior of Cauchy or rescaled variants, but (consistently with intuition) Recentering does not make sense anymore.

\def\bennononenlese{\subsubsection{MLDA testbed TODO new realworld stuff}
Fig. \ref{mldaoneshot} presents results in the MLDA (Machine Learning and Data Analysis) testbed~\cite{mlda}. This testbed is unrelated to the competition won by the Rctg optimization method; the authors of MLDA are not authors of the present study; the parameters of the MetaRctg method have not been tuned on any of these functions; and the functions in MLDA are rooted in the real-world. The original MLDA testbed is not dedicated to one-shot optimization; we use it as a standard tool, now available in the public Nevergrad TODOXX, and have added our one-shot version in that platform for reproducibility~\cite{nevergrad}.
\begin{figure*}
    \centering
    \includegraphics[width=.8\textwidth]{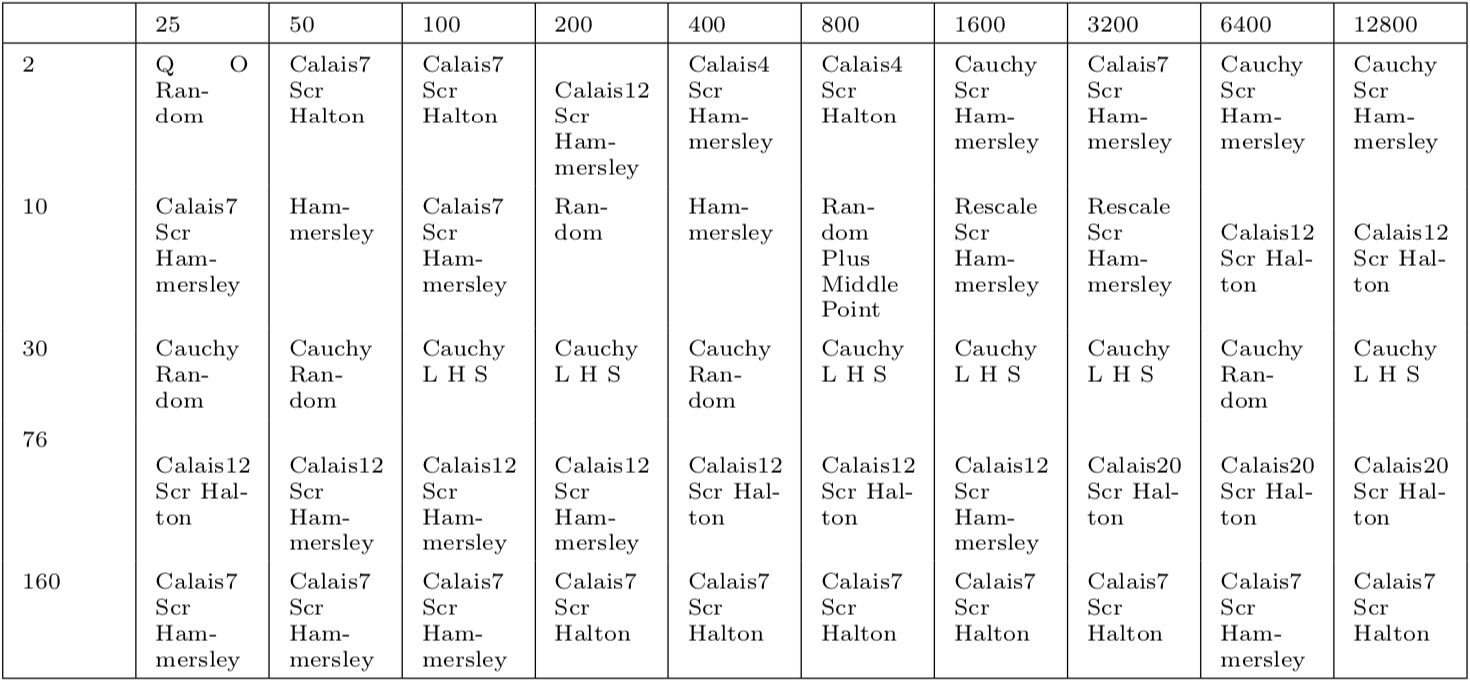}
    \caption{\label{mldaoneshot} Experiments on the MLDA testbed.}
\end{figure*}
 {TODOXX conclusion with metacalais}
}
\section{Conclusion}
\paragraph*{Theoretical results.}
We showed that LHS and jittered sampling have, up to a constant factor, the same stochastic dispersion rate as random sampling (Sections \ref{sec:LHS} and \ref{sec:jittered}), including when restricted to a subset of variables. Low-discrepancy sequences have good stochastic dispersion rates but their performance depends on the order of variables so that they are preserved for subsets of variables only up to a constant which depends on which variables are selected. \ot{Typically} they are outperformed by LHS when there is a single randomly positioned critical variable. Overall, the benefit of sophisticated space-filling designs turn out to be moderate compared to random search or LHS, hence the motivation for alternative fully parallel hyperparameter search ideas such as reshaping.
We then showed that adding a middle point helps in many high-dimensional cases (Section \ref{addmiddle}). This element inspired the design of search distributions different from the Gaussian one, such as Cauchy and/or reshaping as in Rctg methods (Section \ref{calais}) - which provide substantial improvements.
\def\nosf{A drawback of space-filling designs is 
{
that corner-avoidance discussed in~\cite{owencor} and~\cite{othercor} has an impact i.e. weaker results for low-discrepancy sequences compared to random  when the optimum is on a side.
}}

\paragraph*{Practical recommendations.} Table \ref{thomem} surveys our practical conclusions.
{\begin{table}[h]\begin{small}
\ot{\begin{tabular}{|c|c|c|}
\hline
Context & Recommendation & XPs\\
\hline
\hline
$P_0$ perfectly known & MetaCtrng & Table \ref{ngoneshot} \\
\hline
$P_0$ approx. known (rescaled & CchyMetaCtrng & Table \ref{rwoneshot}\\
 data, normalized params) & & Fig  \ref{pgzfig}. \\
\hline
$P_0$ wildly unknown & Cchy- & Table \ref{unscaled}, \\
(avoid this!)& something ?& Fig. \ref{air12}.\\
\hline
\end{tabular}
\caption{Practical recommendations.\label{thomem}}}\end{small}
\end{table}}
Our experiments suggest to 
    {\bf{use MetaRctg (Scrambled-Hammersley + Rctg reshaping with $\lambda$ as in Eq. \ref{metacalais}) when we have a prior on the probability distribution of the optimum}} {(Tab. \ref{ngoneshot} and SM)}. Precisely, with a standard normal prior $P_0$ (use copulas, i.e. multidimensional cumulative distribution functions, for other probability distributions), use 
\begin{equation*}
x_{i,j}=Gcdf\left(\frac{1+\log(n)}{4\log(d)} Gcdf^{-1}(ScrH_{i,j})\right)\\
\end{equation*}
for searching in $[0,1]^{d}$ with budget $n$,
where $ScrH_{i,j}$ is the $j^{th}$ coordinate of the $i^{th}$ point in the Scrambled Hammersley sequence. 
In many cases, however, we do not have such a prior on the position of the optimum: experts provide a correct range of values for most variables, but miss a few ones so that best values are extreme. Unfortunately, low-discrepancy methods are weak, by a dimension-dependent factor, for searching close to boundaries - this is known as the corner avoidance effect\cite{owencor,othercor}. Therefore low-discrepancy methods can then perform worse than random, and Cauchy helps.
A second suggestion is therefore {\bf{to use Cauchy in real world cases}}: we got in Tabs \ref{rwoneshot} and \ref{unscaled} or for AIR (Fig. \ref{air12}) or Pytorch-GAN-zoo (Fig. \ref{pgzfig}), i.e. all our real-world experiments, better performances with Cauchy. Last, {\bf{for real-world problems, rescaling data and parameters and using CauchyMetaRctg looks good}} - with a minimum of standardization, we got good results for CauchyMetaRctg in Fig. \ref{pgzfig} (just using human expertise for rescaling) and Tab. \ref{rwoneshot} (just based on standardizing underlying data, so no real known prior $P_0$). When a proper scaling is impossible, we still get good performance for Cauchy variables but $\lambda$ is impossible to guess so that the best method varies from one case to the other (Table \ref{unscaled}) - Cauchy-LHS and Rescale-Scr-Hammersley being the most stable.
\omitme{
 {{\bf{Overall speed-up by using space-filling designs.}}
Experimental results on the GAN tuning problem confirm results in~\cite{bousquet} in other testbeds. We get the same performance as random sampling but with twice less resources, or equivalently we have probability nearly 2/3 of getting, with a same budget, a better performance with scrambled-Hammersley. {\bf{Center-based sampling: the Rctg sampling and adding a middle point.}} In spite of their simplicity, these methods are a game changer in many high-dimensional cases. MetaRctg is therefore our main recommendation.  Combined with the use of Cauchy distributions, it may also sometimes provide important gains. This strategy also performs well as an initialization of Bayesian optimization. Benefits were less clear as an initialization for differential evolution~\cite{de}.}

{\bf{Changing the distribution.}} TODOXX do we keep this
{A lot of work considers how to be more uniform than independent sampling (e.g. LHS, low discrepancy). However, we can also keep independence and change the distribution and in particular use a distribution which is not invariant by rotation, such as Cauchy.
}

{\bf{Corner avoidance}}~\cite{owencor, othercor} shows a major case in which Hammersley variants can be weaker than random, namely when the optimum is in a corner or a side. However in such a case, all methods should be restarted, so that this weakness is not a big deal. Nevertheless, this is a reminder that checking if the best values are on a side is mandatory and such a case should trigger a rerun.}
%
\def\poubelle{Our results imply that the distribution used for random search should not be equal to the prior on the position of the optimum: a theory for deducing the former from the latter is missing; as an experimental result, we propose:
{\center
\begin{tabular}{|c|}
\hline
Use $x_{i,j}=$ $Gcdf\left(\frac{1+\log(budget)}{4\log(dim)} Gcdf^{-1}(ScrH_{i,j})\right)$\\
\hline
\end{tabular}
}
for searching in $[0,1]^{dimension}$ with budget $budget$,
where $ScrH_{i,j}$ is the $j^{th}$ coordinate of the $i^{th}$ point in the Scrambled Hammersley sequence,
i.e. the Rctg method with $\lambda=\frac{1+\log(budget)}{4\log(dimension)}$. 
We note however that in real world cases, with a less controlled prior on the position of the optimum, hypotheses such as ``the optimum is normally distributed'' rarely occur, and we got our best performance overall with
$x_{i,j}=$ $Gcdf\left(0.55 CauchyCdf^{-1}(ScrH_{i,j})\right)$.
}
}

\FloatBarrier
\setcounter{tocdepth}{14}

\bibliographystyle{apalike} 

\bibliography{doe}
\setcounter{tocdepth}{14}

\end{document}